\pdfoutput=1

\documentclass[11pt]{article}

\usepackage{EMNLP2023}

\usepackage{times}
\usepackage{latexsym}

\usepackage[T1]{fontenc}

\usepackage[utf8]{inputenc}

\usepackage{microtype}

\usepackage{inconsolata}

\usepackage{url}

\usepackage{graphicx}

\usepackage{multirow}
\usepackage{microtype}

\title{Investigating the Encoding of Words in BERT's Neurons using Feature Textualization}

\newcommand{\affilsup}[1]{\rlap{\textsuperscript{\normalfont#1}}}

\author{
    Tanja Baeumel\affilsup{1}
    \qquad
    \textbf{Soniya Vijayakumar}\affilsup{1}
    \qquad
    Josef van Genabith\affilsup{1, 2}
    \\
    \textbf{Guenter Neumann}\affilsup{1, 2}
    \qquad
    \textbf{Simon Ostermann}\affilsup{1}
    \\
    $^1$German Research Center for Artificial Intelligence (DFKI) \\
    $^2$Department of Language Science and Technology, Saarland University \\
    Saarland Informatics Campus, Saarbrücken, Germany\\
    \texttt{\{firstname.lastname\}@dfki.de} \\
}

\begin{document}
\maketitle
\begin{abstract}
Pretrained language models (PLMs) form the basis of most state-of-the-art NLP technologies. Nevertheless, they are essentially black boxes: Humans do not have a clear understanding of what knowledge is encoded in different parts of the models, especially in individual neurons. The situation is different in computer vision, where feature visualization provides a decompositional interpretability technique for neurons of vision models. Activation maximization is used to synthesize inherently interpretable visual representations of the information encoded in individual neurons. 

Our work is inspired by this but presents a cautionary tale on the interpretability of single neurons, based on the first large-scale attempt to adapt activation maximization to NLP, and, more specifically, large PLMs. We propose \textit{feature textualization}, a technique to produce dense representations of neurons in the PLM word embedding space.  We apply feature textualization to the BERT model \cite{devlin-etal-2019-bert} to investigate whether the knowledge encoded in individual neurons can be interpreted and symbolized. 
We find that the produced representations can provide insights about the knowledge encoded in individual neurons, but that individual neurons do not represent clear-cut symbolic units of language such as words.
Additionally, we use feature textualization to investigate how many neurons are needed to encode words in BERT. 

All data and code is made publicly available under \url{https://github.com/BaeumelTanja/Feature-Textualization}.

\end{abstract}
\section{Introduction}

In recent years, research on explainable AI (XAI) has seen an upsurge due to the black box nature of large neural models that are ubiquitously used in state-of-the-art systems. While being highly performant, due to their massive amount of parameters, it is inherently incomprehensible to humans which and how information is stored in these models. Techniques in XAI seek interpretations of large neural models and explanations for their behavior.

\citet{Lipton2016TheMO} identifies post-hoc interpretability and transparency as crucial model properties that enable interpretations. Post-hoc interpretations often do not reveal precisely how a model works, but they can provide useful information for end users. Most local explanation methods fall into this category (see \citet{madsen2022post} for a recent survey on post-hoc interpretations).  
Model transparency on the other hand implies an understanding of how the model works. In this work we are primarily interested in model transparency at the level of \textit{decomposability}, i.e., XAI methods that provide intuitive interpretations for individual neurons.  
In theory, a perfectly transparent and decomposable global interpretation of a neural model could be achieved by providing faithful interpretations of all individual model neurons.

In this work we attempt to find symbolizable and thus intuitive interpretations of individual language model neurons. To obtain interpretations of individual neurons, we build on \textit{feature visualization} \cite{olah2017feature}, an interpretability technique in computer vision that provides visualized, and thus interpretable, representations of individual neurons in vision models. 
In feature visualization, \textit{activation maximization} \cite{erhan2009visualizing} is used to synthesize an input image that maximizes a predefined neuron's activation value. The task is phrased as an optimization problem and gradient ascent is used to iteratively optimize an artificial input with respect to the neuron's activation value. The underlying assumption is that the input image that maximally activates a neuron visualizes the kind of information that is encoded in that neuron. In computer vision, this technique has provided global interpretations of vision models, and has led to important insights: for instance, that certain neurons encode specific patterns such as stripes or textures in a picture \cite{olah2017feature}. The \textit{OpenAI Microscope} \cite{openAImicroscope} is a collection of visualizations of every significant layer and neuron of 13 important vision models, and an impressive demonstration of the relevance of feature visualization in vision models.

In this work we introduce feature \textit{textualization}, an adaptation of activation maximization applicable to language models. Our method provides the - to our knowledge - first attempt at directly interpreting BERT's neurons in a full white box manner without imposing structural constraints. Our contributions with this work are three-fold:
\begin{itemize}
\itemsep0em 
    \item We conduct the - to our knowledge - first large-scale quantitative evaluation of the results of applying activation maximization to large pretrained language models. We present a set of exploratory experiments in which we employ activation maximization to generate input that maximally activates neurons in BERT \cite{devlin-etal-2019-bert}.
    \item We investigate whether feature textualization produces symbolizable interpretations of individual neurons as words.
    To that end, we optimize inputs for single neurons in BERT and compare the resulting vectors to the embeddings of real words. Our findings suggest that the information encoded in single neurons cannot generally be symbolized in terms of words.
    \item  We subsequently investigate how many neurons are required for symbolic interpretations, given the distributedness of information in neural networks. We synthesize artificial inputs for meaningful groups of neurons and find that jointly optimizing an input for 250 to 450 neurons seems to result in vectors that are semantically close to words.
\end{itemize}
\section{Related Work}
\label{sec:related-work}
\paragraph{Interpreting Neurons in PLMs.}
Many previous works have explored interpretation methods for NLP models through the investigation of individual neurons. In an attempt to find concept-level interpretations of individual neurons, multiple studies (e.g., \citet{kadar-etal-2017-representation}, \citet{Na2019DiscoveryON}) pass synthetic n-gram inputs to a model and try to pinpoint the concept that is encoded in the neuron under investigation by automatically extracting a theme across the most activating inputs. This methodology can be useful for neurons that encode multi-word concepts such as phrases \cite{sajjad2022neuron}. However, since the synthetic multi-word inputs are often ungrammatical, there is a risk of identifying a response to arbitrary behavior (like repetition) instead of concept specific behavior \cite{sajjad2022neuron}. Avoiding this obstacle, \citet{bolukbasi2021interpretability} use dataset samples as inputs to detect concepts encoded in individual neurons of BERT. They fall pray to an \textit{interpretability illusion}, as neurons initially seem to encode concepts within datasets, however across datasets these concepts are entirely unrelated. The authors conclude that the knowledge encoded in individual neurons is not reflected by any of the dataset-concepts.
In \citet{Mu2020CompositionalEO}, the neurons to be interpreted are not hand-picked in advance, but the behavior of all neurons is observed and meaningful neurons are chosen depending on their response to certain inputs. The authors compare the presence or absence of different concepts in the input with a binary activation mask on each neuron, i.e., is the neuron more or less activated than a threshold. They use correlations between concepts and neural activations to generate compositional explanations. \citet{Suau2020FindingEI} use neuron activation values as prediction scores for concepts and determine concept knowledge within that neuron via the prediction accuracy. \citet{bills2023language} present an attempt at generating explanations of a language model's behavior using GPT-4 \citep{OpenAI2023GPT4TR}, based on neuron activation patterns. Unlike our work, they train an explainer model to generate explanations, effectively rendering the explanation process as a black-box method again: Neuron activations are not interpreted directly, but based on a second model's prediction. 

To our knowledge, only two other attempts \cite{poerner-etal-2018-interpretable,BERTDream} have previously been made to use activation maximization in the language domain. \citet{poerner-etal-2018-interpretable}  employ activation maximization to synthesize inputs to highly activate single neurons in a small joint vision and language model. Unlike our work, they force synthesized inputs to correspond to one-hot encodings, using the Gumbel-Softmax trick. This induces the strongest possible constraint and bias on the generation process of artificial inputs, as it forces inputs during optimization to converge to words or n-grams of words. This simplification makes it impossible to find an optimal input if it falls in between words or out of the embedding cone completely, a restriction that we lift with our work. \citet{BERTDream} follows a comparable approach to \citet{poerner-etal-2018-interpretable}. \citet{sajjad2022neuron} provides a comprehensive survey on interpretation attempts of individual neurons in NLP.

\paragraph{Activation Maximization in Computer Vision.} Activation maximization was introduced by \citet{erhan2009visualizing}. 
\citet{nguyen2016multifaceted} and \citet{olah2017feature} present applications of feature visualization based on activation maximization in vision models.
Consequently, there has been a large strand of research on finding good regularizers for activation maximization \citep{mahendran2016visualizing,NIPS2016_5d79099f,yosinski-2015-ICML-DL-understanding-neural-networks}. \citet{nguyen2019understanding} provide a comprehensive overview of the use of feature visualization in computer vision.

\section{Feature Textualization}
\label{sec:method}
\subsection{Activation Maximization}
We use activation maximization to generate individual interpretations of individual neurons.
In activation maximization, an input is iteratively optimized to increasingly activate a target neuron. The optimization uses gradient ascent to change the input with respect to the target neuron activation, while keeping all model parameters frozen. To apply activation maximization to the neurons of a model, a continuous optimizable input is required. 

In the standard BERT architecture, inputs, i.e. vector representations of words before being passed to the first BERT layer, are comprised of three parts that are summed: Token embeddings, positional embeddings, and segmentation embeddings. For an input word $w$, let $input_{BERT}()$ be the function that applies these three operations to a one-hot encoding $x_w$ of $w$. The dimensionality of $x_w$ corresponds to the vocabulary size, i.e., $30,522$.

$input_{BERT}(x_w)$ is, in the standard formulation of BERT, a 768-dimensional vector, which is then passed to the first BERT encoder layer.

\begin{figure}[t]
    \centering
    \includegraphics[width=0.2\textwidth]{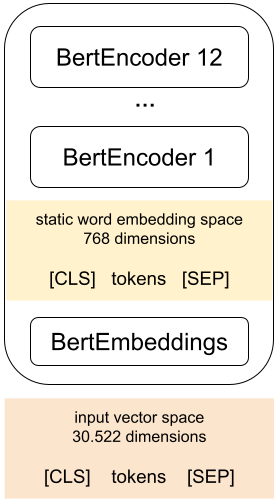}
    \caption{Our analyses are based on the yellow representation after the static embedding layer. We optimize the orange representations in the input vector space.
    }
    \label{fig:BERT}
\end{figure}

In BERT, we optimize a vector at the level of $x_w$, i.e., in the $30,522$-dimensional input vector space, as depicted in the orange part of \autoref{fig:BERT}. We choose to optimize inputs in the  input vector space, because we want to avoid that the activation maximization optimization models positional or segmentation information - it should focus on the meaning representation of the word. Note that unlike e.g. \citet{poerner-etal-2018-interpretable}, we do not force the vectors to converge to actual one-hot encodings, but allow them to take continuous values.

All analyses are then based on the vectors $input_{BERT}(x_w)$ in the 768-dimensional static word embedding space, i.e., after the transformation of $x_w$ by the embedding layer, as depicted in the yellow part of \autoref{fig:BERT}. We interpret the optimized inputs in this static word embedding space, as the input vector space does not encode semantic similarity of its embeddings. In the 768-dimensional static word embedding space, we can compare the optimized input to the embeddings of words in BERT's vocabulary. 

The activation maximization procedure on BERT follows five steps:

\begin{enumerate}
    \setlength{\itemsep}{0.1pt}
    \setlength{\parskip}{0pt}
    \item Select an individual or a group of target neurons for which to optimize an input.
    \item Choose an input length $l$ to be generated, i.e., the number of interpretable token vectors that should be contained in the optimized input.
    \item Create a random initial input in the input vector space, consisting of a
    one-hot encoded [CLS] token, followed by randomly initialized $l$ tokens to be optimized, followed by a one-hot encoded [SEP] token.
    \item Execute a forward pass of the input through BERT. Model weights, and the one-hot inputs at the [CLS] and [SEP] positions are frozen.
    \item Perform gradient ascent with the target neuron activation as the maximization objective. The gradient update is only applied to the $l$ input vectors, i.e. not to the $[CLS]$ and $[SEP]$ one-hot vectors.
\end{enumerate}
We use activation maximization to find optimized inputs in two settings, which are described below: Optimizing for a single neuron, and optimizing for a group of neurons simultaneously.

\subsection{Single Neuron Analysis}
\label{subsec:single-neurons}
We optimize inputs with respect to individual neurons using vanilla gradient ascent without any regularization. In these experiments, we generate inputs to maximally activate individual neurons. For the optimization of a single neuron $i$ we optimize its activation value $a_i$.

\subsection{Groups of Neurons}
\subsubsection{Identifying Meaningful Groups}
\label{subsubsec:determine_neuron_groups}
When optimizing groups of neurons, the first question is which groups of neurons should be optimized together, and how many neurons are needed to encode meaningful and symbolizable units, such as words or concepts. 

We identify such meaningful groups of neurons through a simple, data-driven process: We observe the activation patterns elicited by all words in the vocabulary passed into BERT individually, and store for each target word $w$ the set $I_{w}$ of its $k$ most important neurons. We explore different measures of importance that are elaborated in the next section. We then apply activation maximization to the neuron set $I_{w}$ by maximizing the average activation across $I_{w}$, to generate the optimized input for the group of target neurons.

\subsubsection{Identifying the Most Activated Neurons}
\label{subsubsection:identifying}

We employ two options for computing the set of relevant neurons $I_w$ for a word $w$:

\paragraph{Absolute.} The most simple approach to find the set of most relevant neurons for a word is to calculate for each neuron $i$ the absolute activation value $a_{w,i}^{abs}$ elicited by $w$. We then select the $k$ neurons with the highest values for $a_{w,i}^{abs}$ to compose $I_w$.
\paragraph{Relative:} We choose the $k$ most activated neurons in \textit{relative} terms. Let $a^{max}_i$ be the maximal activation of neuron $i$, determined via the highest activation of neuron $i$ as elicited by the most activating vocabulary item for that neuron. 
We normalize the absolute activation of the neuron $i$ elicited by the word $w$ with the maximal possible activation for $i$: 
\begin{equation}
    a_{w,i}^{rel} = \frac{a^{abs}_{w,i}}{a^{max}_i}
\end{equation}
We choose the $k$ neurons with the highest values $a_{w,i}^{rel}$ for $I_w$ . 

For the optimization of a group of $k$ neurons $I$ and their respective activation values $a_i$, we optimize the objective $\sum_{i \in I}\frac{a_i}{k}$.
\begin{table*}[ht]
\small
\centering
\begin{tabular}{l|l|l}
\hline
\textbf{} & \textbf{Optimized Input} & \textbf{Most activating word}\\
\hline
\textbf{Activation strength} & 42.523 ($\sigma=10.15$) & 1.283 ($\sigma=1.935$)\\
\textbf{cossim($oi_{i}, closest word_{oi}$)} & 0.124 ($\sigma=0.02$) & 1.0 ($\sigma=0$) \\
\textbf{Mean vector magnitude} & 21.818 & 16.867\\
\hline
\end{tabular}
\caption{\label{tab:single-neuron-activation-strength}
Left: Mean activation strength of investigated neurons in response to their respective optimized inputs, cosine similarity of optimized input $oi_i$ to the closest word embedding, mean magnitude of optimized input. Right: same metrics, but based on the most activating word (determined for each neuron individually)
}
\end{table*}

\section{Experiments: Single Neurons}
\label{sec:single}
This Section describes our experiments of applying feature textualization to single neurons in BERT, to determine whether we can find intuitive interpretations through symbolic units of language for these neurons.
\subsection{Experimental Setup}
To understand what type of knowledge is represented in individual neurons, we perform activation maximization on individual neurons. 
In addition to synthesizing an optimized input for neurons, we also get scores on the maximal achievable activation of each neuron, $a_i^{max}$ (s. Section \ref{subsubsection:identifying}).
We run all experiments with single neurons for 5000 optimization steps with a learning rate of 100\footnote{Hyperparameter tuning has revealed such an unusual learning rate to work best, as smaller values do not lead to a convergence of the input generation. We conjecture this is due to the minimal impact a single neuron has on the network.}. Hyperparameter tuning revealed that the activation strength at the neuron under investigation reliably converges within the chosen number of steps.

The neuron activations we maximize are at the level of the output of \textit{the last dense layer per encoder layer}, i.e. before the final layer normalization. For simplicity reasons, we decide to synthesize inputs in the form of a single word, i.e. of a total length of $3$, including the \textit{[CLS]} and \textit{[SEP]} tokens. 

In our experiments, we consider $9,216$ neurons, i.e. the $768$ neurons in all $12$ layers that correspond to the input position of the token to be optimized, i.e., position $1$ in the input. From these, we sample a random 10\% of neuron positions per layer (i.e. 77 out of 768) and optimize the resulting 924 neurons at the respective positions per layer. For all experiments, we use a pretrained and non-finetuned \textit{bert-base-uncased} model from \textit{Hugging Face}\footnote{\url{https://huggingface.co/}}.

\subsection{Evaluation}
\begin{figure}[t]
    \centering
    \includegraphics[width=0.43\textwidth]{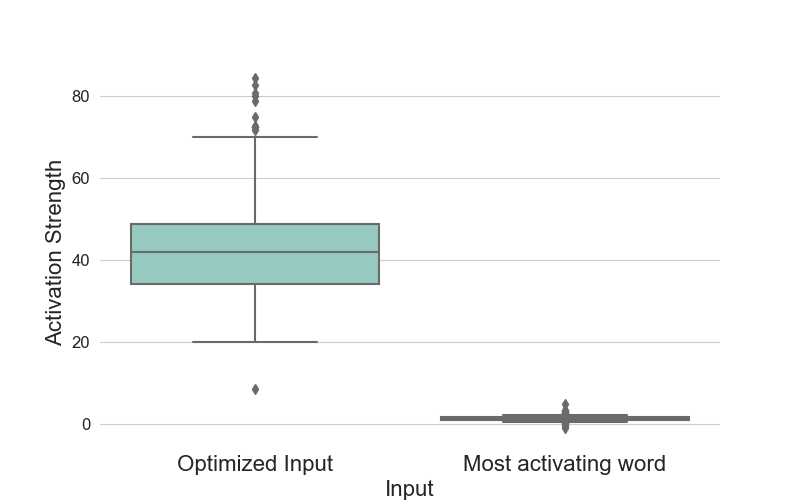}
    \caption{Activation strength for each neuron in response to optimized input and most activating word, averaged over layers}
    \label{fig:Strength}
\end{figure}

\begin{figure}[t]
    \centering
    \includegraphics[width=0.45\textwidth]{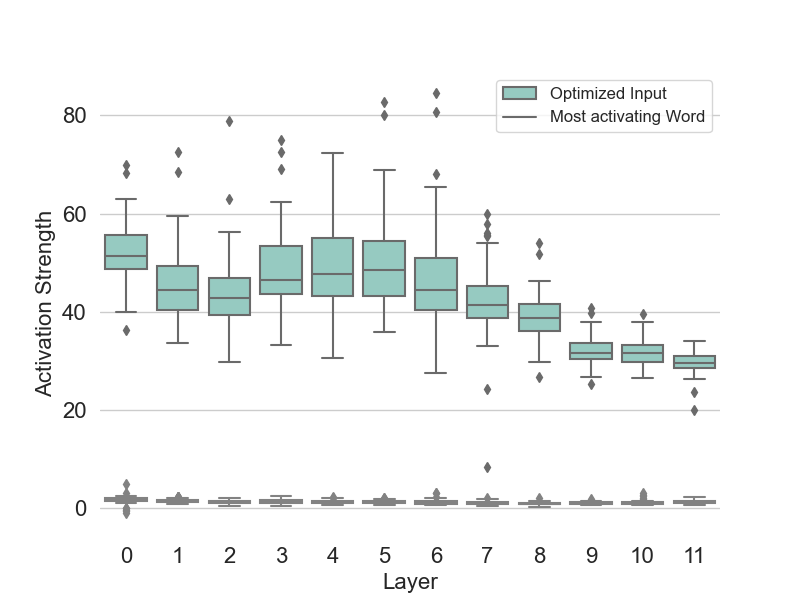}
    \caption{Activation strength for each neurons in response to optimized input and most activating word, per layer}
    \label{fig:StrengthByLayer}
\end{figure}

\paragraph{Activation Potential.}
We compare the 768-dimensional \textit{optimized inputs} in the input embedding space with the embeddings of BERT's input vocabulary in the same space. Specifically, for each neuron that we apply activation maximization to, we also find the most activating vocabulary item (\textit{most activating word}) through a brute-force approach. \autoref{tab:single-neuron-activation-strength} shows quantitative results of our experiments on single neurons. \autoref{fig:Strength} shows that when the gradient ascent optimized input of a specific neuron is used as input, the resulting activation strength at that neuron ($mean = 42.523, \sigma = 10.15$) is more than $30$ times higher than it is when the most activating word for the same neuron is used as input ($mean = 1.283, \sigma = 1.935$). This gives a quantitative indication of how unfaithful single-neuron interpretation methods are when only considering discrete words as possible inputs (as the Gumbel Softmax trick for instance does): By selecting the most activating word as an interpretation of the knowledge encoded in a neuron, only around \textbf{$3\%$} of the actual activation potential are achieved. It is not clear that this can result in a faithful interpretation, as the true knowledge encoded in the neuron is clearly not captured. \\
We find that the activation potential of individual neurons significantly declines with higher layers (\autoref{fig:StrengthByLayer}), as measured by the activation strength achieved by the optimized input ($\beta=-1.87, t(922)=-25.09, p<0.001$). By contrast, the activation strength achieved on neurons based on their respective most activating word remains relatively stable across layers, decreasing only slightly ($\beta=-0.03, t(922)= -8.78, p<0.001$). The difference between word-based and optimization based activation potentials slightly decreases in higher layers, but is still at a level that calls into question the faithfulness of word-based neuron explanations. 

\begin{table*}[!ht]
\centering
\begin{tabular}{l|l|l}
\textbf{Neuron}  & \multicolumn{1}{l|}{\textbf{Layer}}        &   \textbf{top 3 closest words }                                                 \\\hline
0       & \multirow{3}{*}{1}   & triple (0.15), slightest (0.14), serie (0.13)       \\
225     &                             & swollen (0.12), triassic (0.12), skate (0.12)       \\
574     &                             & fuscous (0.12), sicilian (0.12), snails (0.12)      \\ \hline
0       & \multirow{3}{*}{12} & castile (0.16), browser (0.16), U+0F0B (0.15) \\
225     &                             & contradictory (0.19), cerambycidae (0.17), conflicting (0.16)      \\
574    &                             & overheard (0.14), rfc (0.13), sioux (0.13) 
\end{tabular}
\caption{\label{tab:singleexamples}Closest words for the optimized inputs of 3 random neurons, with cosine similarities.}
\end{table*}

\paragraph{Proximity of Optimized Inputs to Words.}
For each optimized input vector, we find the closest vocabulary embedding, through a brute-force approach. We use cosine similarity, since it has a fixed scope\footnote{Cosine similarity is the cosine of the angle between two vectors, and does not depend on the magnitudes of the vectors. The cosine similarity is always in the interval [-1, 1], where two proportional vectors have a cosine similarity of 1, two orthogonal vectors have a similarity of 0, and two opposite vectors have a similarity of -1.} as opposed to for instance Euclidean distance, and is thus easier to interpret. To exclude the possibility that a difference in magnitudes of word embedding vectors and optimized inputs negatively affect our results (vectors with higher magnitude might have more activation potential in general), we compare magnitudes and find that the difference is small (factor 1.29, see \autoref{tab:single-neuron-activation-strength}) in comparison to the difference in activation potential. 

\begin{figure}[t]
    \centering
     \includegraphics[width=0.45\textwidth]{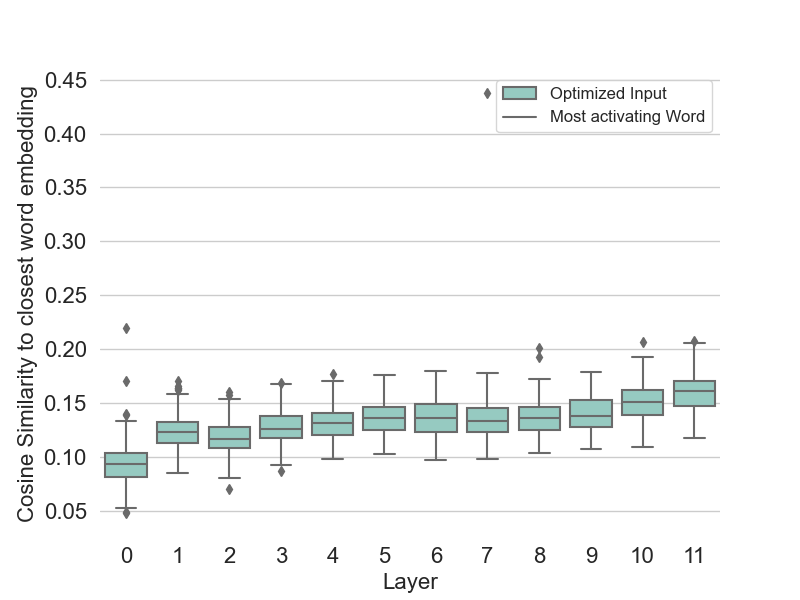}
    \caption{Cosine Similarity of optimized input and closest word embedding}
    \label{fig:CosSim_By_Layer}
\end{figure}

The cosine similarity between an optimized input and the closest embedding of a vocabulary item is on average $0.124$ ($\sigma=0.02$). \autoref{tab:singleexamples} shows the closest words for a number of optimized inputs. \autoref{fig:CosSim_By_Layer} shows that the similarity increases slightly in higher layers ($\beta=0.005, t(2026)=36.73, p<0.001$), such that the mean cosine similarity between an optimized input and its closest word embedding is $0.159$ ($\sigma=0.02$) in the highest BERT layer. 

To understand the scale of typical similarity values in the BERT static embedding space, it is best to consider an example: The similarity of the word \textit{sofa} to \textit{couch} is $0.67$, to \textit{pillows} it is $0.33$ and to \textit{unit} it is $0.15$. In total, only $100$ out of $30,000$ words reach a similarity of over $0.3$ for the \textit{sofa} example. Roughly $50\%$ of words fall into the similarity range of $0.1-0.2$. Thus, a similarity of $0.159$ does not indicate a high semantic relatedness.

Interestingly, the closest word embedding to an optimized input for neuron $i$ does not coincide with the most activating word for the same neuron $i$ for the vast majority of investigated neurons ($99.7\%$). We take this as further evidence on the meaninglessness of interpreting neurons through the most similar word embeddings of their optimized inputs.

\begin{figure}[t]
    \centering
     \includegraphics[width=0.4\textwidth]{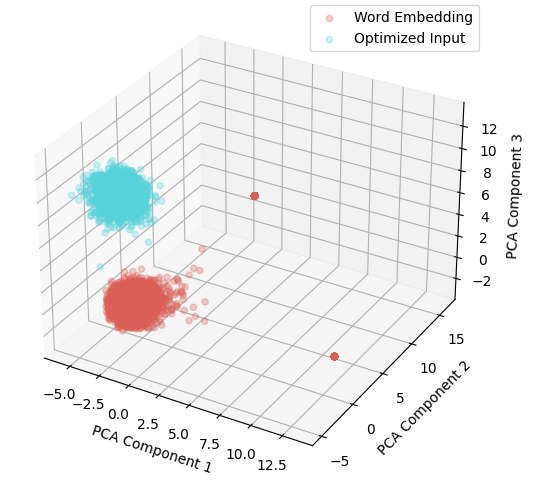}
    \caption{PCA of the space occupied by optimal inputs and words. The two outlier clouds contain non-English symbols and unused tokens and account for approx. $7,000$ instances.}
    \label{fig:PCAWords}
\end{figure}

A visual inspection of the optimized inputs and vocabulary embeddings in the embedding space reveals that optimized inputs and vocabulary embeddings occupy separate parts of space, see \autoref{fig:PCAWords}.

\begin{table*}[t]
\centering  
\begin{tabular}{l|l|l|l|l}
\hline
 & & &  \multicolumn{2}{c}{\textbf{Activation Strength}}\\
\textbf{configuration} & \textbf{k} & \textbf{cossim($oi_{i}$, $w_{i}$}) &  \multicolumn{1}{c}{\textbf{optimized input $oi_{i}$}} &  \multicolumn{1}{c}{\textbf{target word $w_{i}$ }}\\
\hline
\textbf{absolute} & 10 & 0.015 ($\sigma$=0.04) & 13.267 ($\sigma$=7.41) & 1.683 ($\sigma$=0.23)\\
\textbf{absolute} & 100 & 0.140 ($\sigma$=0.06) & 3.320 ($\sigma$=1.10) & 1.071 ($\sigma$=0.15)\\
\textbf{absolute} & 250 & 0.198 ($\sigma$=0.09) & 2.008 ($\sigma$=0.44) & 0.884 ($\sigma$=0.12)\\
\textbf{absolute} & 450 & 0.263 ($\sigma$=0.08) & 1.490 ($\sigma$=0.26) & 0.770 ($\sigma$=0.10)\\
\hline
\textbf{relative} & 10 & 0.008 ($\sigma$=0.05) & 8.943 ($\sigma$=2.39) & 1.072 ($\sigma$=0.21)\\
\textbf{relative} & 100 & 0.085 ($\sigma$=0.06) & 2.403 ($\sigma$=0.53) & 0.872 ($\sigma$=0.12)\\
\textbf{relative} & 250 & 0.207 ($\sigma$=0.09) & 1.480 ($\sigma$=0.28) & 0.771 ($\sigma$=0.10) \\
\textbf{relative} & 450 & 0.266 ($\sigma$=0.06) & 1.160 ($\sigma$=0.19) & 0.694 ($\sigma$=0.09)\\
\hline
\end{tabular}
\caption{\label{tab:groups}
Mean cosine similarity of the target words $w_i$ to the respective optimized input $oi_i$; mean activation strength of the neuron groups and target words, repectively.
}
\end{table*}

\subsection{Interim Conclusion}
We find that:
\begin{itemize}
    \itemsep0em 
    \item Words only utilize the theoretical activation potential of single neurons to a rate of $3\%$ on average, with a slightly higher rate in higher layers.
    \item The average cosine similarity of gradient ascent optimized inputs and their closest words is so small that it most likely does not indicate semantic relatedness.
    \item The words that are closest to optimized inputs do not coincide with the vocabulary items that most strongly activate a neuron.
    \item Words and optimized inputs occupy different subspaces.
\end{itemize}

These results strongly suggest that single neurons do not encode words. Optimized inputs could be dissimilar to word embeddings because the knowledge encoded in single neurons might not be directly symbolizable into words. In fact, this is quite likely, since information in neural networks is known to be distributed.

\section{Experiments: Groups of Neurons}

\label{sec:groups}
In the previous Section, we report evidence suggesting that single neurons do not encode words. Therefore, in this Section, we use activation maximization to try to answer the question: How many neurons does it take to encode words? We present a set of experiments on optimizing an input for \textit{groups} of neurons.

\subsection{Experimental Setup}
We perform activation maximization on groups of neurons to investigate how many neurons are needed to generate an input that is symbolizable into a word. We choose a total of 100 words (henceforth \textit{target words}) of which 80 words are randomly drawn from the vocabulary, disregarding vocabulary items that contain non-latin characters, and an additional 20 words that are manually selected to ensure the presence of high-frequency words in the set. We determine the $k$ most relevant neurons $I_{w}$ for each target word $w$ in absolute and relative terms as described in Section \ref{subsubsection:identifying}, for $k={10, 100, 250, 450}$.
We perform experiments with the same parameters as for the single neuron experiments.

\begin{table}[t]
\centering
\begin{tabular}{l|llll}
\hline
\multicolumn{2}{l}{\textbf{$w_{i}$ is closest word}} & & & \\ \hline
\textbf{k} & \textbf{10} &  \textbf{100} &  \textbf{250} &  \textbf{450} \\
\textbf{absolute} & 1\% & 22\% & 36\% & 53\%\\
\textbf{relative} & 0\% & 48\% & \textbf{67\%} & 65\%\\
\hline \hline
\multicolumn{2}{l}{\textbf{$w_{i}$ is in 20 closest words}} & & & \\ \hline
\textbf{k} & \textbf{10} &  \textbf{100} &  \textbf{250} &  \textbf{450} \\
\textbf{absolute} & 3\% & 32\% & 59\% & 66\%\\
\textbf{relative} & 2\% & 63\% & \textbf{75\%} & 68\%\\
\hline
\end{tabular}
\caption{\label{tab:groups_qualitative}
How often is the target word $w_{i}$ the closest word to the optimized input, or in the top 20?
}
\end{table}

While the selection of neuron groups that encode words based on word activation patterns may seem circular, it is intended to give insights as to whether feature textualization is suited for generating interpretable, word-like inputs when appropriate, as well as insights into the number of neurons $k$ that together encode a word.

\begin{table*}[!t]
\centering
\begin{tabular}{l|l}
\textbf{target word}                    & \textbf{Most similar words}                      \\ \hline
\textbf{hilltop}       & hilltop (0.28), rooftop (0.20), horizon (0.18), buckingham (0.18)        \\\hline
\textbf{corresponding} & corresponding (0.38), correspond (0.27), corresponds (0.23), attached (0.21)         \\\hline
\textbf{crowd}        & crowd (0.38), crowds (0.230), audience (0.23), audiences (0.19)  
\end{tabular}
\caption{\label{tab:groups_examples}
Hand-picked examples of top 4 closest words for optimized inputs of neuron groups, for $k=250$.}
\end{table*}

\subsection{Evaluation}
\paragraph{Neuron Group Size.} We quantitatively evaluate how similar the optimized inputs for the $k$ top activated neurons are to the respective target words.
\autoref{tab:groups} shows that the largest set of optimized neurons ($n=450$) produces optimized inputs that are most similar to the respective target word, both for the relative ($cossim(oi_{i}, w_{i})=0.266$) and the absolute ($cossim(oi_{i}, w_{i})=0.263$) configuration. These somewhat higher similarity values indicate an increased semantic relatedness between groups of neurons and words, as compared to single neurons.

We also find that the most similar vocabulary item to the optimized inputs corresponds to the neuron group's target word in a majority of cases, for sufficiently large groups. \autoref{tab:groups_qualitative} shows that this is the case for 67\% of optimized neuron groups in the \textit{relative, $k=250$} condition. These results also give us a good first intuition as to the number of neurons required to encode a word in BERT: 10 neurons is certainly not enough, and 450 may already be too many. Even though the cosine similarity of the optimized input to the closest word is larger for 450 than for 250 neurons, the word that the optimized input is closest to corresponds to the target word less often ($65\%$ and $68\%$) than in the 250 neuron condition ($67\%$ and $75\%$) as seen in \autoref{tab:groups_qualitative}. Our results could indicate that words are encoded in BERT in neuron clusters of size 250 to 450.

When looking at examples for the closest words to the generated inputs for the top configuration with $k=250$ (s. \autoref{tab:groups_examples}), we find many positive examples which indicate that the selected group of neurons seems to encode a word or concept: Usually, the top most similar words are mutually semantically related. However, for the cases where the most similar word is \textit{not} the target word, we find many cases where the most similar words are random and contain mostly symbols.

\paragraph{Activation Potential.} Similar to the results on individual neurons, the mean activation potential of the optimized input on the optimized set of neurons is higher compared to the target word. While the difference in activation potential between optimized input and most activating word was $33$ times higher for individual neurons, for groups of neurons of size 450 in the relative configuration the activation potential differs only by a factor of $1.7$ (see last row of \autoref{tab:groups}). These findings suggest that words may in fact be suitable interpretations for certain groups of neurons. Note that the mean activation strength decreases for increasingly large groups of neurons, which is to be expected, as small groups of most activated neurons tend to contain neurons with higher activation potential. \\

\section{Discussion}
Our experiments with feature textualization on single neurons indicate that the meaning of single neurons cannot be mapped to words. Unregularized feature textualization resulted in optimized inputs that are dissimilar from all words, but provide maximally faithful interpretations of neurons. We provide a quantitative evaluation of these maximally faithful interpretations compared to word-level interpretations, as obtained in related work. Our results suggest that word-level interpretations of neurons are not faithful and thus not suitable. We thus take our results as a strong contraindication that words could serve as good interpretations of individual neurons in BERT. Especially the dramatic difference in activation potential between optimized inputs and most activating words can be taken as evidence that individual neurons are not particularly sensitive to words and do not directly encode them. 

Our work thus far only investigates the quality of word-level interpretations of LLM neurons. In future work, we plan to investigate what high-level semantic knowledge is encoded within the neurons of an LLM. We are  specifically interested to explore whether individual or small groups of neurons, despite not being interpretable on the word-level, might still be symbolizable in terms of some other form of linguistic or language knowledge.

Our investigation of groups of neurons reveal interesting insights: First, activation maximization and consequently feature textualization does in principle work for language models. The fact that the optimized inputs in the \textit{(relative, $k=250$)} condition are closest to the expected target word in 67\% of cases can be taken as evidence that if a group of neurons truly encodes a word, feature textualization detects that word. We conjecture that whenever the method fails completely in the same condition, i.e., in the 33\% of cases where the target word was not at all similar to the optimized input, the corresponding group of neurons simply did not encode the target word.

A second important insight from our investigations on groups of neurons is that words are certainly not encoded within 10 or less neurons in BERT, as evident by the catastrophic results of all experiments with $k=10$.

In addition to our interpretation of results for single neurons, it is also possible that non-regularized optimal inputs are so dissimilar from words as they are adversarial inputs. Due to the vastness of the 768-dimensional embedding space, it is possible that our procedure finds local maxima that lie in an uninterpretable part of the vector space. This problem is well known in Feature Visualization in Computer Vision: \citet{olah2017feature} find that non-regularized feature visualization produces adversarial examples and thus the use of some regularization or prior is strongly encouraged. 

Based on this, we conjecture that an important avenue for future work is the development of regularizers that pull the optimized inputs further towards the populated part of the word embedding space. We conducted initial experiments with such regularizers and found indications that they help to further increase the similarity of optimal inputs to words and thus increase their interpretability. We will investigate further experiments with regularizers in future work.

The creation of optimal input vectors through activation maximization is associated with a non-neglectable computational cost, since the optimization process is executed for each neuron and group of neurons of interest. The obtained optimal inputs are also model specific. The current method does therefore not scale to a complete investigation of arbitrarily large LLMs. However, recent work by \citet{dar-etal-2023-analyzing} introduces a technique to interpret model parameters in the embedding space in a static way, i.e., with zero passes through the model. Future work should experiment with this approach to investigate the knowledge represented in individual neurons of LLMs. Until then, we make our code and optimized inputs for BERT neurons publicly available, and encourage future work to re-use the published optimal inputs, rather than re-computing them.

Future work also needs to investigate how stable our findings are once the input length is increased, i.e., once words are contextualized. It would also be interesting to apply the method to other language models that are fine-tuned to a downstream task or larger in size.
\section{Conclusion}
In our work, we introduce feature textualization, a decomposability interpretability method for NLP models, based on an adaptation of feature visualization. Our work provides a thorough quantitative evaluation of the application of the method to BERT which leads us to conclude that feature textualization is able to reveal the information encoded in groups of neurons in language models. We find evidence that while single neurons seem to not encode words, feature textualization is able to retrieve interpretable synthetic inputs that are similar to words for well-chosen groups of neurons.
Feature textualization can thus help to pin-point the exact location of encoded knowledge in the parameters of language models.

We hope that our work also shed light on the difficulties found when trying to discretize continuous, dense content representations in language. While the visualization of dense representations in the visual domain is trivial, as arbitrary vectors can be plotted as images, this task is far from trivial in the language domain, where the units that are easiest to interpret are discrete words. We thus see our work as a step towards tackling the problem of making non-discrete representations quasi-interpretable via evaluating their proximity to the interpretable space on a large scale. We think that future work on feature textualization needs to extend such efforts to other units of language, both larger and smaller than words, or other parts of a neural network that go beyond the neuron level.

\section*{Limitations}
The global interpretations that are gained through feature textualization only provide insights into the model that is investigated. Due to the technique requiring optimization for each combination of neurons of interest the method is computationally expensive and consequently infeasible for very large language models. However, since many newer state-of-the-art language models continue to be based on the transformer architecture, interpretations of smaller transformer-based language models, such as BERT, are relevant to this day. To be able to apply feature textualization to these state-of-the-art models, novel zero-pass methods for the interpretation of neurons in the input embedding space need to be developed further. 

\section*{Ethics Statement}
As we do not use any particular sensitive data for our experiments, no specific ethical considerations with respect to the privacy or rights of individuals need to be made. Activation maximization uses gradient ascent and is in principle as computationally expensive as model training, resulting in a significant energy consumption and respective emission of CO$_2$. However, in an effort to limit the energy consumption elicited by this work, we publish the optimized inputs for all neurons, such that these can be reused in future work rather than having to rerun the activation maximization process for the same model.

\section*{Acknowledgements}
We thank the anonymous reviewers for their very valuable feedback. This work has been supported by the German Federal Ministry of Education and Research as part of the project XAINES (01IW20005).

\bibliography{anthology,custom}
\bibliographystyle{acl_natbib}

\end{document}